\documentclass{article}

\usepackage[preprint]{neurips_2023}

\usepackage[utf8]{inputenc} 
\usepackage[T1]{fontenc}    
\usepackage{url}            
\usepackage{booktabs}       
\usepackage{amsfonts}       
\usepackage{nicefrac}       
\usepackage{microtype}      
\usepackage{xcolor}         

\usepackage[pagebackref,breaklinks,colorlinks,bookmarks=false]{hyperref}

\usepackage{natbib}
\setcitestyle{numbers,square,citesep={, }}
\usepackage{graphicx}
\usepackage{multirow}
\usepackage{multicol}
\usepackage{wrapfig}
\usepackage{amsmath,amssymb}
\usepackage{color}
\usepackage{makecell}
\usepackage{colortbl}
\usepackage{xcolor}
\usepackage{pifont}
\newcommand{\cmark}{\ding{51}}%
\newcommand{\xmark}{\ding{53}}%
\newcommand{\eg}{\emph{e.g.}, }
\newcommand{\ie}{\emph{i.e.}, }
\newcommand{\wrt}{\emph{w.r.t.} }

\title{TVTSv2: Learning Out-of-the-box Spatiotemporal Visual Representations at Scale}

\author{%
  Ziyun Zeng$^{1,2}$\thanks{Work done during internship at ARC Lab, Tencent PCG}\quad
  Yixiao Ge$^{2,3 \dagger}$\quad
  Zhan Tong$^3$\quad Xihui Liu$^4$\quad
  Shu-Tao Xia$^{1}$\quad
  Ying Shan$^{2,3}$ \\
  $^1$ Tsinghua University\quad 
  $^2$ ARC Lab, Tencent PCG\quad
  $^3$ Tencent AI Lab\quad  \\
  $^4$ The University of Hong Kong\quad \\
  {$^\dagger$ project lead} \\
}

\begin{document}

\maketitle

\begin{abstract}
The ultimate goal for foundation models is realizing task-agnostic, \ie supporting out-of-the-box usage without task-specific fine-tuning.
Although breakthroughs have been made in natural language processing and image representation learning, it is still challenging for video models to reach it due to the increasing uncertainty of spatiotemporal signals.
To ease training, existing works leverage image foundation models' prior knowledge and equip them with efﬁcient temporal modules.
Despite the satisfactory fine-tuning performance, we empirically find they fall short of out-of-the-box usage, given the even degraded performance in zero-shot/linear protocols compared to their baseline counterparts.
In this work, we analyze the factor that leads to degradation from the perspective of language supervision distortion.
We argue that tuning a text encoder end-to-end, as done in previous work, is suboptimal since it may overfit in terms of styles, thereby losing its original generalization ability to capture the semantics of various language registers.
The overfitted text encoder, in turn, provides a harmful supervision signal, degrading the video representation.
To tackle this issue, we propose a degradation-free pre-training strategy to retain the generalization ability of the text encoder via freezing shallow layers while enabling the task-related semantics capturing in tunable deep layers.
As for the training objective, we adopted the transcript sorting task in TVTS~\cite{TVTS} incorporated with masking techniques~\cite{FLIP} to enable scalable training.
As a result, we produce a series of models, dubbed TVTSv2, with up to one billion parameters.
We achieve new state-of-the-arts on various video benchmarks with a frozen backbone, surpassing the recent ImageBind, InternVideo, \textit{etc}.
Code is available at \url{https://github.com/TencentARC/TVTS}.

\end{abstract}

\section{Introduction}

\begin{figure}[t]
    \centering
    \includegraphics[width=1.0\linewidth]{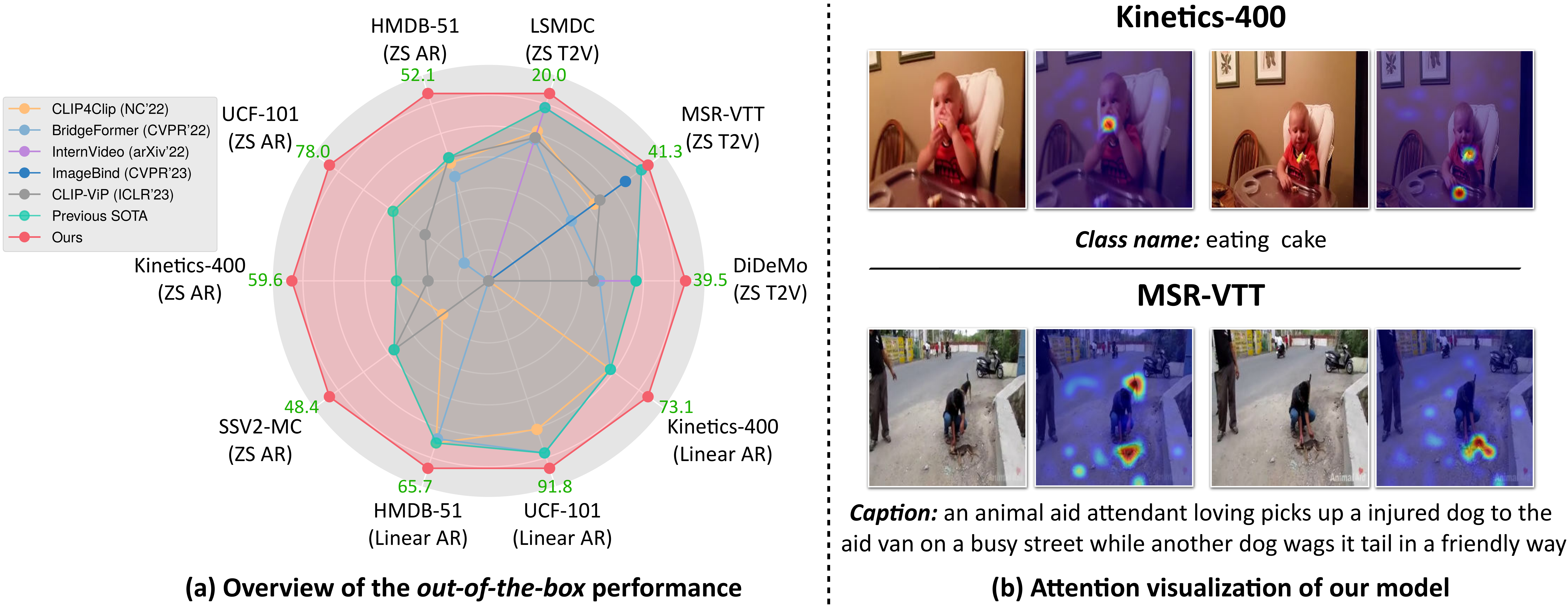}
    \caption{(a) An overview of the out-of-the-box capability of our learned video representations. ZS, AR, and T2V denote zero-shot, action recognition, and text-to-video retrieval, respectively. (b) Visualization of the self-attention distribution of our pre-trained video model. Taking the corresponding class name or caption as a reference, we observe that our video features can well capture the key spatiotemporal context, which explains our good out-of-the-box ability.\vspace{-1em}}
\label{fig:overview}
\end{figure}

Learning universal representations that work out of the box\footnote{The term ``out-of-the-box" indicates that the learned features can be used directly for novel tasks (\eg zero-shot, linear probing) without tailored fine-tuning.} on any downstream task is the ultimate goal for foundation models.
Inspired by the significant success of large language models \cite{gpt1,gpt2,gpt3,gpt4,instructGPT} in natural language processing, a series of efforts have been devoted to migrating this paradigm to computer vision, \eg CLIP \cite{CLIP}, DINO v2 \cite{dinov2}, and ImageBind \cite{imagebind}.
They pre-train visual models with web-crawled or curated data and enable emerging image-centric applications with pre-trained and frozen visual representations, \eg zero-shot image classification, and retrieval.

Despite the progress in learning all-purpose image features, there is still a long way to go in the video domain.
Given the increasing uncertainty of spatiotemporal signals compared to sole spatial ones, it poses a great challenge for learning task-agnostic and universal video representations.
To meet the demand for out-of-the-box usage, a straightforward solution is to scale up the pre-trained video data to cover most distributions. 
Unfortunately, such an idea is in contrast to the fact that the magnitude of publicly accessible videos is much smaller than images, \eg HowTo100M~\cite{howto100m} versus LAION-5B~\cite{laion5b}.
Moreover, the nearly quadratic-increased computational overhead in Transformer~\cite{attention} is generally unaffordable for training video foundation models at a billion scale.

Some works, therefore, turn to exploit the pre-learned spatial prior in image foundation models and adapt them to the video domain by equipping them with temporal reasoning modules, \eg the recent CLIP-ViP~\cite{CLIP-ViP}.
Although such methods achieve state-of-the-art performance when fine-tuned on specific downstream tasks, off-the-shelf video representations are not yet suitable for out-of-the-box usage, exhibiting poor zero-shot/linear results and even degradation from CLIP baselines.
This may be the reason why frame-level CLIP features are still widely used as video feature extractors \cite{clipfeats1,clipfeats2} despite many efforts that have been made in developing video models.
There is a need for real video foundation models that can generate general-purpose video representations.

In this paper, we conduct an in-depth empirical analysis of why previous video models degraded.
Intuitively, the performance degradation mainly comes from the overfitting on the relatively small-scale post-pretrain data, sacrificing the generalization ability of the original foundation models.
To this end, many existing works focused on efficient visual tuning with adapters~\cite{st-adapter,aim}, while ignoring the potentially overfitted text encoder, which produces distorted text supervision that could, in turn, degrades video representations.

Different from the sole caption knowledge learned from the image foundation models, video data introduces ASR transcripts, which contain temporal dependency and facilitate temporal reasoning, and have been widely adopted in recent literature~\cite{CLIP-ViP,TVTS} to boost spatiotemporal learning.
Such supervision provides valuable temporal information but shows a great domain gap from the pre-learned alt-text.
Our extensive experiments reveal that performance degradation in prior methods stems from the text encoder's compromised capabilities of generalizing to various language styles 
due to the end-to-end fine-tuning with noisy ASR transcripts. This not only hinders the proper semantic capture behind the ASR transcripts but also impairs pre-learned language knowledge within the text model, thus negatively affecting video representation learning.

Given the observations, we propose a degradation-free pre-training strategy with a partially frozen text encoder, \textit{i.e.}, the shallow layers are frozen while the deep layers are tunable.
With such a strategy, the pre-trained text encoder can generalize well to various language registers with different styles, \textit{e.g.}, ASR transcripts for training, and search queries for downstream retrieval.
The new language semantics and structures can be captured in deep layers, thus producing semantically meaningful learning targets to advance out-of-the-box video representation learning.
Regarding the training objective, we adopt the pretext task of Turning to Video for Transcript Sorting (TVTS)~\cite{TVTS} for its superior performance.
To further scale up training to pursue state-of-the-art performance, we adopt the masking strategy without reconstruction~\cite{FLIP} to enable affordable training with larger backbone architectures.
As a result, we successfully trained huge-size models with one billion parameters in total using 80 V100 GPUs in one week.

Finally, we offer a series of pre-trained models, dubbed TVTSv2, from base size to huge size. Compared to the original TVTS \cite{TVTS}, we inherit the rich semantic knowledge learned from CLIP-pretrained models and scale up the models by up to 7 times. The knowledge inheritance is actually not trivial given the degradation of prior methods \cite{CLIP-ViP,actionclip} as we discussed above.
As illustrated in Figure~\ref{fig:overview} (a), our pre-trained model produces all-purpose spatiotemporal visual representations, that can be used for zero-shot/linear video classification, and zero-shot video-text retrieval on various datasets out of the box.
It is well noticeable that our TVTSv2 surpasses the recent SOTAs, \ie ImageBind \cite{imagebind} and InternVideo \cite{internvideo}, on zero-shot video classification and retrieval, despite more data or more modalities they leveraged. 
Surprisingly, we also achieve comparable performance to DINOv2-g \cite{dino} on linear K400 with 40\% fewer parameters.
The encouraging results shed light on the direction of developing general-purpose video foundation models.

\section{Related Works}
\textbf{Out-of-the-box Video Representations.} 
In out-of-the-box image representation learning, the supervision signal may come from web-crawled images~\cite{dinov2}, descriptive texts~\cite{CLIP,ALIGN}, or other modalities~\cite{imagebind}, where the second one dominates the literature.
Similarly, in the video domain, a bunch of works has made an effort to shift such a paradigm to pursue out-of-the-box video representations.
For instance, TVTS~\cite{TVTS} adopts a dual-stream architecture 
and learns fine-grained spatiotemporal representation by resorting to videos for transcript sorting.
CLIP4Clip~\cite{CLIP4Clip} stacks a temporal Transformer on top of the origin CLIP to aggregate frame representations.
CLIP-ViP~\cite{CLIP-ViP} plugs several video proxy tokens that attend to different frames for temporal summarization.
InternVideo~\cite{internvideo} inherits UniFormerV2~\cite{uniformerv2} and stacks a cross-modal decoder to enable delicate video-text interaction.
Other works either focus on improving vision signals~\cite{videoclip} or accessing more modalities~\cite{imagebind}.
However, their improvement in zero-shot and linear probe evaluation is marginal, indicating there is still a long way to reach general-purpose video representation.

\textbf{Domain Specialists for Video Tasks.} 
Besides the video pre-training, another line of research adapted the pre-trained foundation models to the specific video tasks for obtaining domain experts.
They reach in-domain gains in two technical routes:
(\textbf{i}) Designing proper objectives.
For instance, Pace~\cite{wang2020self} and SVT~\cite{svt} learn invariant spatiotemporal characteristics, \eg motion correspondences of different objects, by aligning clips sampled in a different frame rate. 
(\textbf{ii}) Parameter-efficient tuning.
For example, ST-adapter~\cite{st-adapter} and AIM~\cite{aim} plug several tunable spatial and temporal adapters into each attention block, leaving the original parameters frozen.
Similarly, visual prompts~\cite{ept,vpt} replace the manually constructed prompt~\cite{CLIP} with learnable ones to raise instance-specific representations.
Nevertheless, such a paradigm is opposite to general-purpose video representation learning, making them inflexible in novel scenarios for real-world applications.

\textbf{Scalable Visual Pre-training.} 
Plenty of work are devoted to improving the model scalability in two aspects:
(\textbf{i}) Scale-up data.
In the video domain, some work harvests from alt-text-video pairs~\cite{frozen,webvidqa} due to the high-quality descriptions, while the available data is limited.
The follow-up works utilize ASR transcripts derived from the raw video~\cite{merlot,howto100m,HD-VILA-100M} for better scalability.
In this work, we reuse the transcript sorting objective~\cite{TVTS} for scalable spatiotemporal learning.
(\textbf{ii}) Reducing computational overheads.
The video pre-training is restricted to the quadratically-increased self-attention complexity so far.
Masked visual modeling is proposed to improve the efficiency of pre-training, which drops a large portion of visual tokens and reconstructs them given unmasked ones.
Although solid results have been reached under a high masking ratio~\cite{mae-st,videomae,videomaev2}, these works are short of out-of-the-box transferability due to train-test mismatch.
Recently, FLIP~\cite{FLIP} directly conducts image-text contrastive pre-training based on masked visual tokens and achieves favorable results.
Inspired by this work, we train the models with up to one billion parameters by properly incorporating masking.

\section{Method}

\begin{figure}[t]
    \centering
    \includegraphics[width=1.0\linewidth]{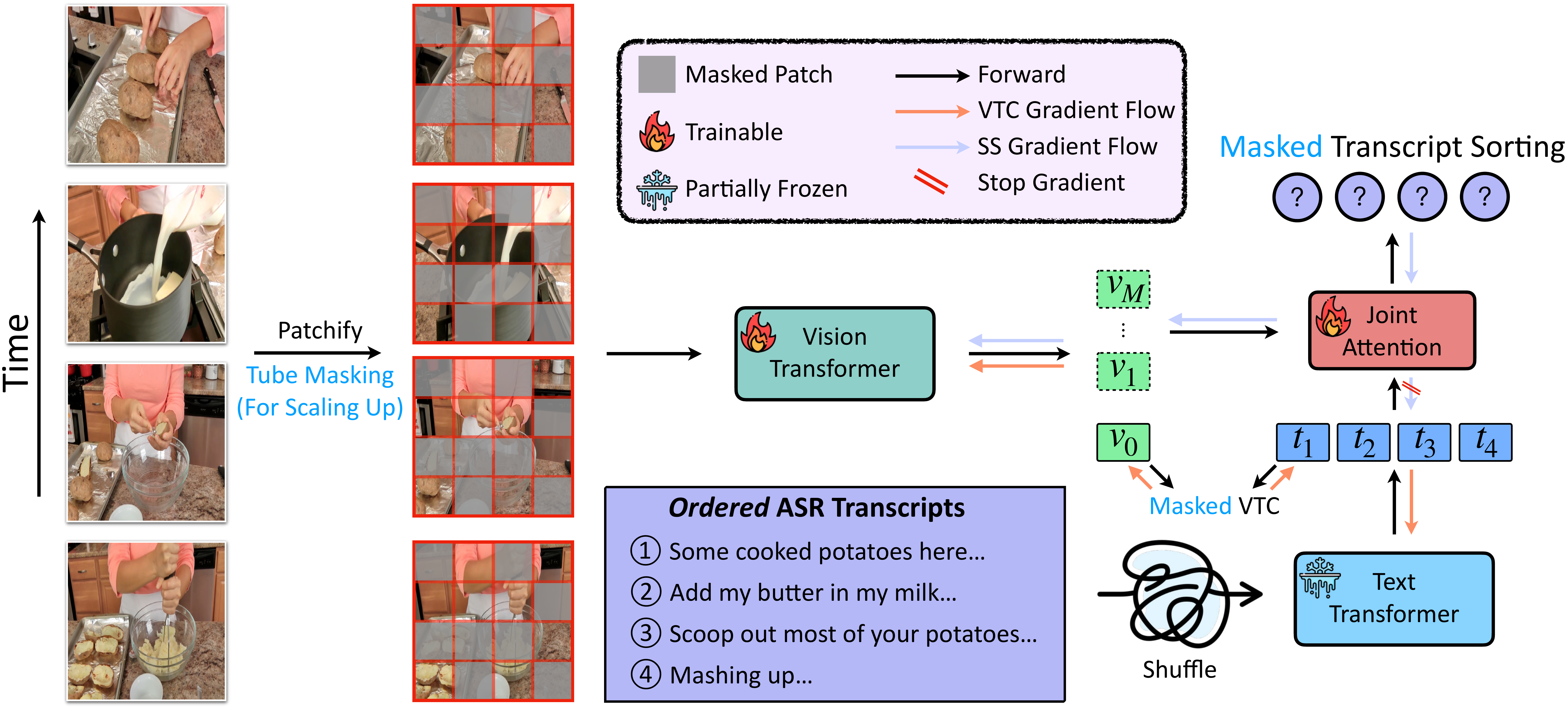}
    \caption{Our training framework.
    A large portion of patches from sampled frames is first dropped out via random tube masking before being sent into a Vision Transformer equipped with divided space-time attention for encoding video representations.
    The corresponding ASR transcripts are shuffled and embedded by a partially frozen text encoder.
    The disposable joint attention (only for training) is performed among all video and text representations for predicting the transcripts' chronological order, formulated by a $K$-way classification objective.
    \vspace{-1em}}	
\label{fig:framework}
\end{figure}

In this section, we introduce how to learn out-of-the-box video representations at scale without the performance degradation of the image pre-trained knowledge.
Our main framework is illustrated in Figure~\ref{fig:framework}, where the trained Vision Transformer is deployed for extracting video features that work out of the box on downstream tasks.
We will introduce our model architectures in Section~\ref{subsec:model_arch}, our partially frozen training strategy to avoid knowledge degradation in Section~\ref{subsec:partially_frozen}, our masking mechanism to enable scalable pre-training in Section~\ref{subsec:masking_strategy},
and our training objectives in Section~\ref{subsec:training_objectives}.

\subsection{Model Architectures}
\label{subsec:model_arch}
To inherit the knowledge of pre-trained image foundation models, \ie CLIP, we apply the dual-stream framework, which consists of a visual encoder and a text encoder. We extend the visual encoder to the video domain by equipping it with divided space-time attention following the practice in \cite{timesformer}.

\textbf{Video Encoder.}
We adopt the standard Vision Transformer (ViT)~\cite{vit} with $L_V$ layers as the video encoder and add divided space-time attention~\cite{timesformer} in each layer for spatiotemporal reasoning.
Given $T$ frames sampled from a video with a resolution of $3\times H\times W$, we split each frame into patches of size $P\times P$, then project them to a token sequence $v \in\mathbb{R}^{TN \times D_V}$, where $N=HW/P^2$ and $v_i\in\mathbb{R}^{D_V}$ denotes the \emph{i}-th token with a dimension $D_V$.
Next, learnable spatial and temporal positional embeddings, denoted as $\mathbf{E}^s\in\mathbb{R}^{N\times D_V}$ and $\mathbf{E}^t\in\mathbb{R}^{T\times D_V}$, are added to each token, \ie
\begin{equation}
    v_{x,y}^{(0)} = v_{x,y} + \mathbf{E}_{x}^{s} + \mathbf{E}_{y}^{t}
\end{equation}
where $v_{x,y}$ denotes the \emph{x}-th token sampled from the \emph{y}-th frame.
All \emph{x}-th tokens sampled from different frames are given the same spatial positional embedding, and all tokens belonging to the \emph{y}-th frame are given the same temporal positional embedding.
Finally, a learnable [CLS] token $v_0\in\mathbb{R}^{D_V}$ is concatenated at the beginning of the sequence, and we perform divided space-time attention across the $L_V$ layers. 
The [CLS] token output by the $L_V$-th layer is further projected to a $D$-dimensional shared space and used as our final video representation for out-of-the-box usage.

\textbf{Text Encoder.}
The text encoder contains $L_T$ stacked Transformer blocks.
Following CLIP~\cite{CLIP}, we adopt the casual attention mechanism.
Similar to the visual branch, we project the [CLS] token $t_0\in\mathbb{R}^{D_T}$ output by the last layer to the $D$-dimensional shared space as the final text representation.

\subsection{Degradation-free Pre-Training with Partially Frozen Text Encoder}
\label{subsec:partially_frozen}

Learning video representations with the assistance of text dominates recent literature~\cite{frozen,bridgeformer,CLIP-ViP}.
There are two main lines of research:
(\textbf{i}) Learning video features under alt-text supervision, \textit{i.e.}, video caption. The alt-text is generally clean but hard to scale up, \eg WebVid-10M \cite{frozen}.
(\textbf{ii}) Using the ASR transcripts as supervision which are naturally tied with the video and easy to scale up, \eg YT-Temporal~\cite{merlot}. Despite some noise, transcripts generally provide temporal dependency that facilitates temporal reasoning. 

Existing methods using only alt-text \cite{frozen,bridgeformer,MILES} are proven to improve video tasks marginally due to the limited amount of such data. Subsequent works \cite{TVTS,CLIP-ViP} have focused more on the proper utilization of large-scale ASR transcripts in order to improve the ability of temporal modeling on downstream tasks. Despite the impressive results after fine-tuning, their out-of-the-box video representations are actually degraded compared to the pre-trained image counterparts, \textit{i.e.}, the zero-shot capability is inferior to simply aggregating the frame-level image features. 
The recent CLIP-ViP~\cite{CLIP-ViP} argued that it is mainly caused by the domain gap between ASR transcripts and downstream captions, and proposed to use auxiliary captions generated by OFA~\cite{OFA} for training. Unfortunately, the generated captions actually do not work for out-of-the-box video representation learning. Our detailed empirical analysis is as follows.

\begin{wraptable}{r}{8cm}
\setlength{\tabcolsep}{3pt}
\centering
\vspace{-1.5em}
\caption{The Mean Matched Similarity (MMS), Recall@1 (R@1) and top-K (K=1, 5) accuracy of different models. The tests are done with the officially released ViT-B/32 models. M$_1$ is trained with YT-Temporal, and M$_2$ is jointly trained with YT-Temporal and WebVid-2.5M.}
\resizebox{1.\linewidth}{!}{
\begin{tabular}{cccccc}
\toprule
\multirow{2}{*}{Method} & \multirow{2}{*}{Text Corpus} &
\multicolumn{2}{c}{DiDeMo} & \multicolumn{2}{c}{Kinetics-400} \\
\cmidrule{3-6}
                        &                              & MMS           & R@1        & Top-1           & Top-5        \\
\midrule
CLIP                    & -                     & 0.295        & 24.7       & 64.5        & 86.7       \\
CLIP-ViP                & ASR + generated-alt-text               & 0.221        & 22.6       & -        & -       \\
M$_1$-FT            & ASR                          & 0.215        & 19.3       & 61.7        & 84.9       \\
M$_2$-FT            & ASR + alt-text               & 0.234        & 24.2       & 64.0        & 86.1       \\
M$_2$-FF   & ASR + alt-text  & 0.187 & 21.9 & 63.5 & 85.9 \\
M$_2$-PF                    & ASR + alt-text               & \textbf{0.340}        & \textbf{29.8}       & \textbf{67.1}        & \textbf{88.2}   \\
\bottomrule
\end{tabular}}
\vspace{-1em}
\label{tab:mms}
\end{wraptable}

\paragraph{An empirical study of video degradation.}
As shown in Table~\ref{tab:mms}, 
we test the out-of-the-box capability of the learned video representations by measuring the video-text Mean Matched Similarity and text-to-video Recall@1 on DiDeMo~\cite{DiDeMo}, and Top-1/5 linearly probing accuracies on Kinetics-400~\cite{kinetics-400}.

As illustrated in L1-2 of Table~\ref{tab:mms}, we observe that the CLIP-ViP~\cite{CLIP-ViP} trained with OFA-generated captions still faces performance degradation compared to the CLIP baseline.
In order to further distinguish the effects of the noisy generated captions from the domain gap, we first build a baseline initialized from CLIP-ViT-B/32, namely \textbf{M$_1$-FT}, and end-to-end optimize the visual and text models with Video-Text Contrastive solely.
The results in Table~\ref{tab:mms} show more degradation than CLIP-ViP.
Next, we add an alt-text training source, \ie WebVid2.5M~\cite{frozen}, denoted as \textbf{M$_2$-FT}, and slightly outperforms CLIP-ViP because the text quality is better than OFA-generated ones.
However, \textbf{M$_2$-FT} still lags behind the baseline, indicating that \textit{simply reducing the domain gap has minor improvements but cannot solve the degradation problem}.

Since CLIP has reached the highest MMS and R@1 so far, we assume that the text knowledge learned from alt-text-image pairs is somewhat suitable for zero-shot inference in the video domain.
The end-to-end tuning paradigm is suboptimal and incurs catastrophic forgetting for the text branch, which in turn provides distorted supervision for the visual branch and hurts out-of-the-box ability.
To verify our assumption, we retrain \textbf{M$_2$-FT} with a fully frozen text encoder, denoted as \textbf{M$_2$-FF}.
However, it performs worse, indicating that the pre-trained and frozen text encoder fails to correctly capture the language semantics and produce harmful learning targets.

Given the above observations, we come up with a simple yet effective training strategy, namely partially frozen (PF), which freezes the shallow layers of the text encoder.
Such a strategy promotes the text encoder's generalizability \wrt different language registers with varying styles.
The semantics and structures of ASR transcripts are captured in deep layers, producing semantically meaningful supervision signals to facilitate out-of-the-box video representation learning.
In our practice, we freeze the first three-quarters of layers, leaving other layers trainable.
As shown in Table~\ref{tab:mms}, the model that adopts the partially frozen training strategy, \ie \textbf{M$_2$-PF}, yields a significant improvement over the baseline.
\emph{Pre-training with a partially frozen text encoder could well preserve the knowledge of the CLIP-pretrained model and unleash it for learning strong out-of-the-box video representations.}

\subsection{Scalable Pre-training with Masked Video Encoder}
\label{subsec:masking_strategy}

We aim to pursue scalable video representation learning with larger backbone architectures in order to achieve better out-of-the-box capabilities.
However, the token length, \ie $T*N$, is always a bottleneck for scaling up video training due to spatiotemporal attention's drastically increased computational budget.
The recent VideoMAEs~\cite{videomae,videomaev2} improve training efficiency by masking a high proportion of patches, significantly reducing the token length.
A similar practice was observed in image pre-training, \ie FLIP~\cite{FLIP}.
Inspired by their success, we experiment with video pre-training at high mask ratios without reconstruction.

Specifically, we employed the tube masking approach introduced by VideoMAE~\cite{videomae} because it's compatible with divided space-time attention and can effectively diminish temporal redundancy. 
Essentially, this method randomly blocks $\rho$\% of patches from the same position across various frames, resulting in a token length of $TN(1-\rho)$. 
For this study, we set $\rho\ge 50\%$, translating to at least half of the computational budget being reduced.
This ensures our approach can be scaled to models with billions of parameters. 

\subsection{Training Objectives}
\label{subsec:training_objectives}

\noindent \textbf{Transcript Sorting (TS).}
\label{subsubsec:TS}
Sorting the unordered transcripts given ordered video patches has been proven effective for learning sophisticated spatiotemporal interaction~\cite{TVTS}.
In this work, we further reveal the scalability of TS for training large models under a high mask ratio.
Given translated words and their timestamps $\{w_i,a_i\}_{i=1}^{N_w}$, where $w_i$ denotes the \emph{i}-th word, $a_i$ denotes its corresponding timestamp (in seconds), and $N_w$ denotes the word number, we first sample $K$ transcript segments $\{T_k\}_{k=1}^K$ of length $l$ (in seconds) with an interval of 1s between consecutive segments:
\begin{equation}
    \begin{split}
    &L_k = L_{\text{start}} + (k-1)*(l+1) \\
    &T_k = \{w_i | a_i \in [L_k,L_k+l]\}
    \end{split}
\end{equation}
where $L_{\text{start}}$ denotes a randomly picked starting time, and $L_k$ denotes the beginning time of the \emph{k}-th segments. 
Then we randomly shuffle the $K$ segments to $\{T_{o_i}\}_{i=1}^K$, where $T_{o_i}$ denotes the $i$-th segment in the shuffled sequence corresponding to the ground-truth chronological order $o_i$. As for frame sampling, we follow TSN~\cite{TSN} to divide the overall interval, \ie $[L_1, L_K+l]$, into $T$ equal-space intervals and randomly sample 1 frame per interval.

After encoding sampled transcript segments and frames, we concatenate the text [CLS] tokens $\{t_0^i\}_{i=1}^K$ with all unmasked video tokens $\{v_i\}_{i=0}^{TN\rho}$ and perform joint attention across them.
Next, we send the attend text [CLS] tokens into a $K$-way classifier and predict their orders separately.
Finally, the transcript sorting objective is formulated as a cross-entropy:
\begin{equation}
    \mathcal{L}_{\text{TS}} = \frac{1}{K}\sum_{i=1}^K -\log \frac{\exp(p_{o_i}^i)}{\sum_{j=1}^K \exp(p_j^i)}
\end{equation}
where $p^i \in\mathbb{R}^K$ denotes the prediction for the \emph{i}-th transcript segment in the shuffled sequence, $p_j^i$ indicates the probability that its chronological order is $j$, and $o_i$ represents the ground-truth order.

It is worth noting that the model may learn shortcuts, \eg natural language order, without attending to visual information.
We prevent such cases from hurting training by stopping gradients flowing to the text encoder, which forces the video model to provide well-learned spatiotemporal context for TS.

\noindent \textbf{Video-Text Contrastive.}
\label{subsubsec:VTC}
We adopt the widely-used Video-Text Contrastive (VTC) as the basic objective for semantic alignment, which is formulated as:
\begin{equation}
    \begin{split}
    &\mathcal{L}_{\text{VTC}} = \text{NCE}(\overline{t},v_0) + \text{NCE}(v_0,\overline{t}) \\
    s.t.\quad &\text{NCE}(q,k) = - \log \frac{\exp(q^Tk_+/\tau)}{\sum_{i=1}^B \exp(q^Tk_i/\tau)}
    \end{split}
\end{equation}
where $\tau$ denotes the temperature parameter, $\overline{t}$ denotes the averaged text [CLS] token, \ie $\overline{t}=\frac{1}{K}\sum_{i=1}^K t_0^i$, and $v_0$ denotes the video [CLS] token. 
Note that $v_0$ only attends to the unmasked patches during encoding.
Our overall training objective is $\mathcal{L} = \mathcal{L}_{\text{VTC}} + \lambda \mathcal{L}_{\text{TS}}$,
where $\lambda$ is a hyperparameter.
\section{Experiments}

\begin{table}[t]
\setlength{\tabcolsep}{3pt}
\centering
\caption{The \textbf{zero-shot text-to-video retrieval} results. $\dagger$ denotes using post-processing DSL~\cite{dsl}. $*$ means accessing extra modalities, \eg audio. The underlined number indicates absolute SOTA. Single-stream models are de-emphasized.}
\resizebox{1.\textwidth}{!}{
\begin{tabular}{llcccc|cccc|cccc}
\toprule
\multirow{2}{*}{Method} & \multirow{2}{*}{Venue} & \multicolumn{4}{c}{MSR-VTT} & \multicolumn{4}{c}{DiDeMo} & \multicolumn{4}{c}{LSMDC} \\
\cmidrule{3-14}
                        &                        & R@1   & R@5   & R@10 & MdR  & R@1   & R@5   & R@10 & MdR & R@1  & R@5  & R@10 & MdR  \\
\hline
\emph{\textcolor{blue}{Non-CLIP models}}         &                        &       &       &      &      &       &       &      &     &      &      &      &      \\
VideoCLIP~\cite{videoclip}               & EMNLP'21               & 10.4  & 22.2  & 30.0 & -    & 16.6  & 46.9  & -    & -   & -    & -    & -    & -    \\
Frozen~\cite{frozen}                  & ICCV'21                & 18.7  & 39.5  & 51.6 & 10.0 & 21.1  & 46.0  & 56.2 & 7.0 & 9.3  & 22.0 & 30.1 & 51.0 \\
ALPRO~\cite{alpro}                   & CVPR'22                & 24.1  & 44.7  & 55.4 & -    & 23.8  & 47.3  & 57.9 & -   & -    & -    & -    & -    \\
\textcolor{gray}{VIOLET}~\cite{violet}                  & \textcolor{gray}{arXiv'22}               & \textcolor{gray}{25.9}  & \textcolor{gray}{49.5}  & \textcolor{gray}{59.7} & \textcolor{gray}{-}    & \textcolor{gray}{23.5}  & \textcolor{gray}{49.8}  & \textcolor{gray}{59.8} & \textcolor{gray}{-}   & \textcolor{gray}{-}    & \textcolor{gray}{-}    & \textcolor{gray}{-}    & \textcolor{gray}{-}    \\
BridgeFormer~\cite{bridgeformer}            & CVPR'22                & 26.0  & 46.4  & 56.4 & 7.0  & 25.6  & 50.6  & 61.1 & 5.0 & 12.2 & 25.9 & 32.2 & 42.0 \\
\textcolor{gray}{OmniVL}~\cite{omnivl}                  & \textcolor{gray}{NeurIPS'22}                & \textcolor{gray}{34.6}  & \textcolor{gray}{58.4}  & \textcolor{gray}{66.6} & \textcolor{gray}{-}    & \textcolor{gray}{33.3}  & \textcolor{gray}{58.7}  & \textcolor{gray}{68.5} & \textcolor{gray}{-}   & \textcolor{gray}{-}    & \textcolor{gray}{-}    & \textcolor{gray}{-}    & \textcolor{gray}{-}    \\
\hline
\emph{\textcolor{blue}{CLIP-B/32}}               &                        &       &       &      &      &       &       &      &     &      &      &      &      \\
CLIP~\cite{CLIP}                    & ICML'21                & 30.6  & 54.4  & 64.3 & 4.0  & 24.7  & 49.3  & 60.9 & 6.0 & 13.6 & 27.9 & 35.5 & 32.0 \\
CLIP-straight~\cite{clipstraight}           & MCPR'21                & 31.2  & 53.7  & 64.2 & 4.0  & -     & -     & -    & -   & 11.3 & 22.7 & 29.2 & 56.5 \\
CLIP4Clip~\cite{CLIP4Clip}               & NC'22      & 32.0  & 57.0  & 66.9 & 4.0  & -     & -     & -    & -   & 15.1 & 28.5 & 36.4 & 28.0 \\
BridgeFormer~\cite{bridgeformer}       & CVPR'22                & 33.2  & 58.0  & 68.6 & 4.0  & -     & -     & -    & -   & 15.5 & 30.7 & 38.7 & 22.0 \\
CLIP-ViP~\cite{CLIP-ViP}                & ICLR'23                & 29.0  & 51.2  & 61.3 & 5.0  & 22.6  & 43.9  & 56.4 & 7.0 & 11.3 & 25.3 & 31.3 & 38.0 \\
\rowcolor[RGB]{207,234,241} Ours-B/32               & -                      & 34.5  & \textbf{58.5}  & 67.7 & 3.5  & 31.2  & 56.9  & 68.3 & 4.0 & \textbf{16.1} & 30.6 & \textbf{38.7} & 25.0 \\
\rowcolor[RGB]{207,234,241} Ours-B/32$^\dagger$         & -                      & \textbf{36.4}  & 58.0  & \textbf{69.0} & \textbf{3.0}  & \textbf{37.0}  & \textbf{61.6}  & \textbf{70.9} & \textbf{3.0} & 15.6 & \textbf{31.8} & 38.6 & \textbf{22.0} \\
\hline
\emph{\textcolor{blue}{CLIP-B/16}}               &                        &       &       &      &      &       &       &      &     &      &      &      &      \\
CLIP~\cite{CLIP}                    & ICML'21                & 31.8  & 53.9  & 64.5 & 4.0  & 27.7  & 51.0  & 62.5 & 5.0 & 15.2 & 29.7 & 37.6 & 25.0 \\
CLIP-ViP~\cite{CLIP-ViP}                & ICLR'23                & 31.7  & 53.8  & 63.2 & 4.0  & 24.6  & 50.7  & 59.7 & 5.0 & 12.5 & 26.1 & 33.3 & 39.0 \\
\textcolor{gray}{UMT-B}~\cite{umt} & \textcolor{gray}{arXiv'23} & \textcolor{gray}{29.6} & \textcolor{gray}{52.8} & \textcolor{gray}{61.9} & \textcolor{gray}{-} & \textcolor{gray}{33.4} & \textcolor{gray}{58.3} & \textcolor{gray}{67.0} & \textcolor{gray}{-} & \textcolor{gray}{16.8} & \textcolor{gray}{30.5} & \textcolor{gray}{37.6} & \textcolor{gray}{-} \\

\rowcolor[RGB]{207,234,241} Ours-B/16               & -                      & 35.9  & 61.2  & 71.3 & 3.0  & 33.4  & 60.1  & 70.6 & 3.0 & 16.9 & 31.5 & 38.2 & 22.0 \\
\rowcolor[RGB]{207,234,241} Ours-B/16$^\dagger$         & -                      & \textbf{37.8}  & \textbf{62.9}  & \textbf{72.4} & \textbf{3.0}  & \textbf{39.0}  & \textbf{63.9}  & \textbf{72.6} & \textbf{3.0} & \textbf{18.3} & \textbf{33.7} & \textbf{41.9} & \textbf{19.0} \\
\hline
\emph{\textcolor{blue}{Larger Models}}           &                        &       &       &      &      &       &       &      &     &      &      &      &      \\
ImageBind$^*$~\cite{imagebind} & CVPR'23 & 36.8 & 61.8 & 70.0 & - & - & - & - & - & - & - & - & - \\
\textcolor{gray}{InternVideo} & \textcolor{gray}{arXiv'22} & \textcolor{gray}{40.0} & \textcolor{gray}{65.3} & \textcolor{gray}{74.1} & \textcolor{gray}{2.0} & \textcolor{gray}{31.5} & \textcolor{gray}{57.6} & \textcolor{gray}{68.2} & \textcolor{gray}{3.0} & \textcolor{gray}{17.6} & \textcolor{gray}{32.4} & \textcolor{gray}{40.2} & \textcolor{gray}{23.0} \\
\textcolor{gray}{UMT-L} & \textcolor{gray}{arXiv'23} & \textcolor{gray}{33.3} & \textcolor{gray}{58.1} & \textcolor{gray}{66.7} & \textcolor{gray}{-} & \textcolor{gray}{34.0} & \textcolor{gray}{60.4} & \textcolor{gray}{68.7} & \textcolor{gray}{-} & \textcolor{gray}{20.0} & \textcolor{gray}{37.2} & \textcolor{gray}{43.7} & \textcolor{gray}{-} \\
\rowcolor[RGB]{207,234,241} Ours-H/14               & -                      & 38.2  & 62.4  & 73.2 & 3.0  & 34.6  & 61.9  & 71.5 & 3.0 & 17.3 & 32.5 & 41.4 & 20.0 \\
\rowcolor[RGB]{207,234,241} Ours-H/14$^\dagger$         & -                      & \underline{\textbf{41.3}}  & \underline{\textbf{63.0}}  & \underline{\textbf{74.0}} & \underline{\textbf{2.0}}  & \underline{\textbf{39.5}}  & \underline{\textbf{63.6}}  & \underline{\textbf{73.1}} & \underline{\textbf{3.0}} & \underline{\textbf{20.0}} & \underline{\textbf{37.8}} & \underline{\textbf{48.6}} & \underline{\textbf{11.0}} \\
\bottomrule
\end{tabular}}
\vspace{-1em}
\label{tab:main_t2v}
\end{table}

\begin{table}[t]
\setlength{\tabcolsep}{2pt}
\centering
\caption{The \textbf{zero-shot action recognition} results of top-1 accuracy. The underlined number indicates absolute SOTA. Single-stream models are de-emphasized. ``Our impl.'' denotes we use official pre-trained weights for evaluation.}
\resizebox{.75\textwidth}{!}{
\begin{tabular}{llcccccc}
\toprule
Method              & Venue & Cite From           & Params                           & HMDB-51 & UCF-101 & Kinetics-400 & SSV2-MC \\
\midrule
\emph{\textcolor{blue}{Non-CLIP models}} &  &  &                                 &         &        &              &         \\
MTE~\cite{mte}                 & ECCV'16 & X-CLIP        & \multirow{10}{*}{\rotatebox{90}{< 200M}} & 19.7    & 15.8   & -            & -       \\
ASR~\cite{asr}                 & ECML'17 & X-CLIP        &                                  & 21.8    & 24.4   & -            & -       \\
ZSECOC~\cite{zsecoc}              & CVPR'17 & X-CLIP        &                                  & 22.6    & 15.1   & -            & -       \\
UR~\cite{ur}                  & CVPR'18  & X-CLIP       &                                  & 24.4    & 17.5   & -            & -       \\
TS-GCN~\cite{ts-gcn}              & AAAI'19  & X-CLIP        &                                  & 23.2    & 34.2   & -            & -       \\
E2E~\cite{e2e}                 & CVPR'20  & X-CLIP       &                                  & 32.7    & 48.0   & -            & -       \\
ER-ZSAR~\cite{er-zsar}             & ICCV'21 & X-CLIP         &  & 35.3    & 51.8   & -            & -       \\
\textcolor{gray}{ClipBert}~\cite{clipbert}            & \textcolor{gray}{CVPR'21} & \textcolor{gray}{BridgeFormer}        &                                  & \textcolor{gray}{20.0}    & \textcolor{gray}{27.5}   & \textcolor{gray}{-}            & \textcolor{gray}{-}       \\
Frozen~\cite{frozen}              & ICCV'21   & BridgeFormer     &  & 27.5    & 45.4   & -            & -       \\
BridgeFormer~\cite{bridgeformer}        & CVPR'22  & BridgeFormer        &                                  & 38.0    & 51.1   & -            & -       \\
\midrule
\emph{\textcolor{blue}{CLIP-B/16}}           &                 &                                  &         &        &              &         \\
CLIP~\cite{CLIP}                & ICML'21 & Our impl.         & \multirow{5}{*}{\rotatebox{90}{< 200M}} & 43.2    & 68.9   & 48.0         & 29.6    \\
ActionCLIP~\cite{actionclip}          & arXiv'21  & X-CLIP      &                                  & 40.8    & 58.3   & -            & -       \\
\textcolor{gray}{X-CLIP}~\cite{X-CLIP}              & \textcolor{gray}{ECCV'22} & \textcolor{gray}{X-CLIP}         &                                  & \textcolor{gray}{44.6}    & \textcolor{gray}{72.0}   & \textcolor{gray}{-}            & \textcolor{gray}{-}       \\
CLIP-ViP~\cite{CLIP-ViP}            & ICLR'23 & Our impl.         &                                  & 41.2    & 58.9   & 37.6         & 35.5    \\
\rowcolor[RGB]{207,234,241} Ours-B/16           & -    & -           &                                  & \textbf{50.4}    & \textbf{69.8}   & \textbf{54.3}         & \textbf{42.1}    \\
\midrule
\emph{\textcolor{blue}{Larger Models}} &   &                                  &         &        &              &         \\
ImageBind~\cite{imagebind} & CVPR'23 &  ImageBind &  \makecell[c]{632M (V) \\ 354M (T)}  & - & - & 50.0 & - \\
\textcolor{gray}{X-Florence}~\cite{X-CLIP}        & \textcolor{gray}{ECCV'22}  & \textcolor{gray}{X-CLIP}       & \textcolor{gray}{\makecell[c]{637M (V) \\ 256M (T)}}                             & \textcolor{gray}{48.4}    & \textcolor{gray}{73.2}   & \textcolor{gray}{-}            & \textcolor{gray}{-}       \\
\rowcolor[RGB]{207,234,241} Ours-H/14           & -  & -              & \makecell[c]{632M (V) \\ 354M (T)}                             & \underline{\textbf{52.1}}    & \underline{\textbf{78.0}}   & \underline{\textbf{59.6}}         & \underline{\textbf{48.4}}    \\
\bottomrule
\end{tabular}}
\vspace{-1em}
\label{tab:main_ar}
\end{table}

\begin{table}[t]
\setlength{\tabcolsep}{2pt}
\centering
\caption{The \textbf{linear action recognition} results of top-1 accuracy. We de-emphasize DINOv2~\cite{dinov2} because it uses a larger model than ours, \ie ViT-g/14. ``Our impl.'' denotes we use official pre-trained weights for evaluation.}
\resizebox{.75\textwidth}{!}{
\begin{tabular}{llcccccc}
\toprule
Method                   & Venue & Cite From     & Supervision  & Params             & HMDB-51 & UCF-101 & Kinetics-400 \\
\midrule
MemDPC~\cite{memdpc}                   & ECCV'20 & SVT   & \multirow{10}{*}{\rotatebox{90}{Self}} & \multirow{9}{*}{\rotatebox{90}{< 100M}} & 30.5    & 54.1    & -            \\
CoCLR~\cite{coclr}                    & NeurIPS'20 & SVT &                         & & 52.4    & 77.8    & -            \\
Vi$^2$CLR~\cite{vi2clr} & ICCV'21    & SVT & &                           & 47.3    & 75.4    & 63.4         \\
VideoMoCo~\cite{videomoco}                & CVPR'21  & SVT  & &                            & 49.2    & 78.7    & -            \\
CVRL~\cite{cvrl} & CVPR'21  & SVT  & &                            & 57.3    & 89.2    & 67.6         \\
DINO~\cite{dino}                     & ICCV'21 & DINOv2    & &                       & -       & 85.0    & 64.5         \\
SVT~\cite{svt}                      & CVPR'22 & SVT   & &                          & 57.8    & 90.8    & 68.1         \\
iBOT~\cite{ibot}                     & ICLR'22  & DINOv2   & &                          & -       & 88.6    & 72.6         \\
VideoMAE-B~\cite{videomae}               & NeurIPS'22 & Our impl. & &                       & 30.9    & 52.7    & 20.4         \\
\textcolor{gray}{DINOv2-g}~\cite{dinov2}                & \textcolor{gray}{arXiv'23}   & \textcolor{gray}{DINOv2} & &    \textcolor{gray}{1.0B}      &         \textcolor{gray}{-}       & \textcolor{gray}{91.2}    & \textcolor{gray}{78.4}         \\
VideoMAEv2-H~\cite{videomaev2}            & CVPR'23  & Our impl.  &  &                    632M       & 34.1    & 56.4    & 25.8         \\
\midrule
CLIP-B/16~\cite{CLIP} & ICML'21 & Our impl. & \multirow{6}{*}{\rotatebox{90}{Language}} & & 62.8 & 87.6 & 66.9 \\
Frozen~\cite{frozen}                   & ICCV'21  & Our impl.  &  & \multirow{6}{*}{\rotatebox{90}{< 100M}} &57.8    & 88.7    & 62.9         \\
MERLOT~\cite{merlot}                   & NeurIPS'21 & MERLOT &  &                         & 49.6    & 74.9    & -            \\
BridgeFormer~\cite{bridgeformer}                     & CVPR'22    & Our impl. & &                          & 60.7    & 89.2    & 65.6         \\
MILES~\cite{MILES}                    & CVPR'22  & Our impl.  &  &                         & 60.0    & 89.6    & 64.0         \\
TVTS~\cite{TVTS} & CVPR'23 & Our impl. & & & 60.1 & 87.6 & 60.8 \\
\rowcolor[RGB]{207,234,241} Ours-B/16                & -  & -        &  &                         & 64.7    & 90.0    & 70.1         \\
\rowcolor[RGB]{207,234,241} Ours-H/14                & -  & -        &  & 632M                        & \textbf{65.7}    & \textbf{91.8}    & \textbf{73.1}        \\
\bottomrule
\end{tabular}}
\vspace{-1em}
\label{tab:main_linear}
\end{table}

\subsection{Experimental Setup}
\noindent \textbf{Pre-training Datasets.}
We jointly pre-train our model on two datasets:
(a) \textbf{YT-Temporal}~\cite{merlot} contains 6M YouTube videos with ASR transcribed words and timestamps.
(b) \textbf{WebVid-2.5M}~\cite{frozen} contains 2.5M alt-text-video pairs. 
Since the timestamps are unavailable, we only perform VTC on it.

\noindent \textbf{Text-to-Video Retrieval.}
We evaluate \emph{zero-shot} performance of our model on three benchmarks:
(a) \textbf{MSR-VTT}~\cite{msrvtt}.
(b) \textbf{DiDeMo}~\cite{DiDeMo}.
(c) \textbf{LSMDC}~\cite{lsmdc}.
The Recall@K (R@K) and Median Rank (MdR) are reported as the evaluation metric.

\noindent \textbf{Action Recognition.}
Four benchmarks are used in this task:
(a) \textbf{HMDB-51}~\cite{hmdb51},
(b) \textbf{UCF-101}~\cite{ucf101},
(c) \textbf{Kinetics-400}~\cite{kinetics-400},
(d) \textbf{SSV2}~\cite{ssv2}.
The evaluation is two-fold:
(\textbf{i}) \emph{Zero-shot Action Recognition}. 
Following~\cite{CLIP-ViP}, we use the prompt template ``a person [CLASS]'' for the first three datasets and evaluate SSV2 on a multi-choice setting, namely \textbf{SSV2-MC}. See Appendix for details.
(\textbf{ii}) \emph{Linear Classification}, where we optimize a linear classifier added on top of the frozen visual encoder. 

\subsection{Implementation Details}
We inherit weights from CLIP~\cite{CLIP} for the standard ViT modules and initialize the parameters of time attention with zeros~\cite{timesformer}.
The temperature parameter in $\mathcal{L}_{\text{VTC}}$ is set to be 0.05, and we sample $K=4$ transcript segments for the transcript sorting task.
As for pre-training and downstream evaluation, $T=12$ frames are sampled with an input resolution of $224\times 224$.
Note that the joint attention for sorting is duplicated in downstream tasks.
The masking ratio $\rho$ is set to $50\%$ for ViT-B/16 and $70\%$ for ViT-H/14, and we do not mask patches for ViT-B/32.
Most hyper-parameters are shared across different models, and we list the training details in Appendix.

\subsection{Main Results}
\noindent \textbf{Text-to-Video Retrieval.}
The zero-shot retrieval results are reported in Table~\ref{tab:main_t2v}.
Compared to previous dual-stream SOTA, our model reaches an absolute gain of $1.4\%$, $6.9\%$ and $1.8\%$ in terms of R@1 on MSR-VTT, DiDeMo, and LSMDC, respectively.
Using post-processing techniques like DSL~\cite{dsl} further boosts performance.
It is worth noting that we even surpass or achieve comparable performance with single stream models, \eg InternVideo~\cite{internvideo}, where the latter has a much higher retrieval complexity and does not own out-of-the-box ability for extracting video features alone.
Moreover, the solid result further verifies the effectiveness of our proposed degradation-free pre-training strategy, especially given the fact that CLIP-ViP~\cite{CLIP-ViP} degrades and CLIP4Clip~\cite{CLIP4Clip} only makes marginal improvement. 

\noindent \textbf{Action Recognition.}
We report the zero-shot action recognition results in Table~\ref{tab:main_ar}.
Our model brings significant improvements, \ie $8.9\%$, $9.1\%$, $11.6\%$, and $12.9\%$ absolute gain on HMDB-51, UCF-101, Kinetics-400, and SSV2-MC, respectively.
The prompt-based method, \ie ActionCLIP~\cite{actionclip}, also degrades on this test, possibly due to overfitting to manually constructed templates. 
When it turns to larger models, we even surpass X-Florecnce~\cite{X-CLIP}, a single-stream model with comparable video encoder parameters, by a large margin, revealing the superior scalability of our training paradigm.
Even for the more challenging benchmark that contains various motion dynamics
, \ie SSV2-MC, our model still achieves promising results, solidifying the contribution of masked transcript sorting that promotes fine-grained spatiotemporal representation learning.

In addition, we provide the liner classification results in Table~\ref{tab:main_linear}, where we surpass either  self-supervised or language-guided models.
Notably, we find the linear accuracy of VideoMAE~\cite{videomae} and VideoMAEv2~\cite{videomaev2} lags far behind other models.
It implies that representation learned by MAE-style objectives concentrates on the pixel level instead of the semantic level, making them impractical for learning general-purpose features.
By contrast, with the aid of language supervision, our model is capable of out-of-the-box transferring and retains training efficiency like MAE variants.

\subsection{Ablation Study}
Considering efficiency, all ablations are based on the ViT-B/32 model if no specified.

\begin{wraptable}{r}{8cm}
\vspace{-1.5em}
\setlength{\tabcolsep}{3pt}
\centering
\caption{The zero-shot text-to-video retrieval results \wrt different objectives. ``sg'' denotes stopping gradients.}
\resizebox{1.\linewidth}{!}{
\begin{tabular}{llllcccccc}
\toprule
\multirow{2}{*}{Name} & \multirow{2}{*}{$\mathcal{L}_{\text{VTC}}$} & \multirow{2}{*}{$\mathcal{L}_{\text{TS}}$} & \multirow{2}{*}{sg} & \multicolumn{2}{c}{MSR-VTT} & \multicolumn{2}{c}{DiDeMo} & \multicolumn{2}{c}{LSMDC} \\
\cmidrule{5-10}
                      &                         &                        &                     & R@1          & R@5          & R@1          & R@5         & R@1         & R@5         \\
\midrule
CLIP                  & n/a                       & n/a                      & n/a                   & 30.6         & 54.4         & 24.7         & 49.3        & 13.6        & 27.9        \\
M$_\text{base}$               & \cmark                       & \xmark                      & n/a                   & 33.0         & 56.7         & 29.8         & 55.5        & 15.2        & 29.3        \\
M$_\text{w/o sg}$             & \cmark                       & \cmark                      & \xmark                   & 31.9         & 56.2         & 29.2         & 54.9        & 14.9        & 28.9        \\
M$_\text{ours}$               & \cmark                       & \cmark                      & \cmark                   & 34.5         & \textbf{58.5}         & 31.2         & 56.9        & 16.1        & 30.6        \\
M$_\text{ours}$++             & \cmark                       & \cmark                      & \cmark                   & \textbf{35.0}         & 58.4         & \textbf{31.7}         & \textbf{57.3}        & \textbf{16.5}        & \textbf{31.2}       \\
\bottomrule
\end{tabular}}
\vspace{-1em}
\label{tab:ab_obj}
\end{wraptable}

\noindent \textbf{Training Objectives.}
We first analyze the impact of each training objective in Table~\ref{tab:ab_obj} and reach the following conclusions:
(\textbf{i}) M$_\text{base}$ outperforms the CLIP baseline, implying that both ASR transcripts and alt-texts are reliable supervision that benefits video representation learning.
(\textbf{ii}) M$_\text{ours}$ outperforms $M_{\text{base}}$, while M$_\text{w/o sg}$ underperforms.
It indicates that without stopping gradients flowing to the text encoder, the model turns to learn shortcuts, in other words, optimize the transcript representation to ease the sorting task.
The wrong optimization direction harms training efficiency.
Instead, M$_\text{ours}$ encourages enhancing spatiotemporal representation to provide enough knowledge for transcript sorting, remarkably improving the performance.

Note that we cannot perform TS on WebVid2.5M because the timestamps are unavailable.
In this case, we further generalize TS from transcript sorting to text sorting by concatenating different clips and their corresponding alt-texts in a batch, then predicting the alt-text order, namely M$_\text{ours}$++.
Such a naive trial boosts the performance to some extent, revealing the potential of ``resorting to video for text sorting'' in facilitating spatiotemporal representation learning.

\begin{wrapfigure}{r}{8cm}
\vspace{-1.2em}
    \centering
    \includegraphics[width=1.\linewidth]{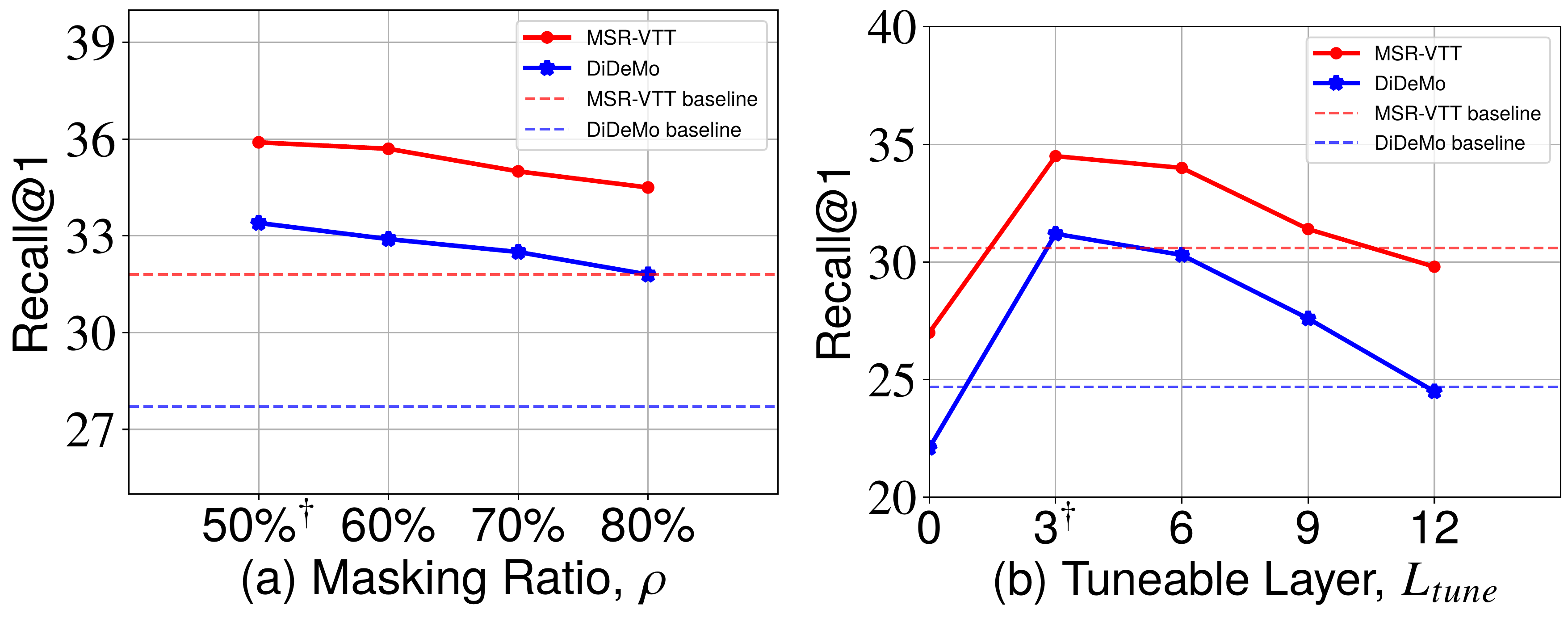} \\
    \vspace{-0.5em}
    \caption{The sensitivity of (a) masking ratio $\rho$, (b) tuneable layer $L_{\text{tune}}$. Default settings are marked with $\dagger$.}	
\vspace{-0.8em}
\label{fig:ab_other}
\end{wrapfigure}

\noindent \textbf{Masking Ratio.}
We report the Recall@1 under different masking ratio $\rho$ in Figure~\ref{fig:ab_other} (a) based on the ViT-B/16 model. 
We empirically find that, given the same training steps, smaller masking ratios lead to better performance due to the more learned tokens.
We imply that over-masking, \eg $\rho=80\%$ is too hard for the model to capture enough visual semantics despite more training efficiency.
As a result, for different scales of models, we need to find a proper masking ratio that could trade off the performance and training cost.

\noindent \textbf{Tuneable Layer.}
The performance in terms of different tuneable layers $L_\text{tune}$ (count from the last layer) for the text encoder is reported in Figure~\ref{fig:ab_other} (b).
$L_\text{tune}=0$ equals to fully freeze the text encoder, and $L_\text{tune}=12$ means fully fine-tuning.
Both $L_\text{tune}=0$ and $L_\text{tune}=12$ lag far behind partially freezing the text encoder, verifying the effectiveness of the proposed degradation-free pre-training strategy, which preserves the image foundation model's pre-learned knowledge while unleashing it to learn powerful out-of-the-box spatiotemporal representation.

\vspace{-1em}
\section{Conclusion}
\vspace{-0.5em}

In this work, we pursue out-of-the-box spatiotemporal visual representations with a newly proposed TVTSv2.
Compared to TVTS~\cite{TVTS}, we introduce a degradation-free pre-training strategy to solve the representation degradation from the pre-trained image foundation model, which is actually non-trivial given the failure of previous models~\cite{CLIP-ViP,actionclip}.
We further adopt the masking technique \cite{FLIP} to improve scalability.
With TVTSv2, we train a series of models with up to one billion parameters, achieving state-of-the-art results in terms of zero-shot and linear probe evaluation on various video tasks.
Notably, our model even surpasses or is competitive with those trained on more data or modality, \eg InternVideo~\cite{internvideo} and ImageBind~\cite{imagebind}.
In a nutshell, we make a step towards learning out-of-the-box spatiotemporal visual representation despite some limitations: (i) Though performance improvements on the public benchmark, the emergent abilities are not present yet. (ii) The largest model studied in this paper is still far from the SOTA model in the image domain (\textit{i.e.}, ViT-22B \cite{dehghani2023scaling}).

{\small
\bibliographystyle{ieee_fullname}
\bibliography{egbib}
}

\appendix
\section{Downstream Tasks}
\noindent \textbf{Text-to-Video Retrieval.}
The statistics of the three zero-shot text-to-video retrieval benchmarks are listed as follows: 
(a) \textbf{MSR-VTT}~\cite{msrvtt} consists of 10K videos harvested from YouTube, and there are around 200K descriptions. 
We follow prior works~\cite{CLIP4Clip,CLIP-ViP,internvideo} to conduct evaluations on the 1K-A test set.
(b) \textbf{DiDeMo}~\cite{DiDeMo} consists of 10K Flickr videos with around 40K sentences.
Following~\cite{CLIP4Clip,bridgeformer,frozen}, we concatenate all sentences that describe the same video to form a single query and conduct paragraph-to-video retrieval. 
Specifically, we do not crop and concatenate the localized moments but directly use the whole video in the retrieval set, as done by~\cite{bridgeformer,frozen}.
(c) \textbf{LSMDC}~\cite{lsmdc} has 118,081 videos cropped from 202 movies. 
The evaluation protocol follows~\cite{TVTS,bridgeformer}, where the test set contains 1,000 videos. 

\noindent \textbf{Action Recognition.}
The statistics of the four zero-shot/linear action recognition benchmarks are listed as follows:
(a) \textbf{HMDB-51}~\cite{hmdb51} contains 5K videos of 51 action categories. 
The training and test sets have 3.5K and 1.5K videos, respectively.
(b) \textbf{UCF-101}~\cite{ucf101} contains 13K videos of 101 action categories. 
The training set has 9.5K videos, and the test set has 3.5K videos.
(c) \textbf{Kinetics-400}~\cite{kinetics-400} is a large-scale dataset with 260K videos belonging to 400 categories, where 240K videos are used for training, and 20K videos are used for validation.
(d) \textbf{SSV2}~\cite{ssv2} consists of 189K videos showing humans performing 174 pre-defined fine-grained actions with everyday objects.
The training set has 169K videos, and the validation set contains 20K videos.

In zero-shot action recognition, we follow the prior work~\cite{CLIP} to use the prompt template “a person [CLASS]” for \textbf{HMDB-51}, \textbf{UCF-101}, and \textbf{Kinetics-400}, where the cosine similarity is calculated between all video-category pairs for classification.
As for \textbf{SSV2}, we turn it into a multi-choice task, namely \textbf{SSV2-MC}.
For each video, we randomly pick 173 negative descriptions from other categories (one description per category) and put them along with the ground-truth one into a candidate set. 
The model is expected to retrieve the right one from the 174 candidates.

\section{Implementation Details}

For our ViT-B/32 and ViT-B/16 models, we load the weights released by OpenAI~\cite{CLIP}.
As for the largest ViT-H/14 model, we inherit weights from OpenCLIP~\cite{laion5b}.
The detailed model architectures are listed in Table~\ref{tab:model_arch}.

We use AdamW~\cite{adamw} as the optimizer with a weight decay of $0.05$.
The initial learning rates are set to be $1\times 10^{-4}$ and $1\times 10^{-7}$ for the newly added modules and the origin CLIP modules, respectively.
We fix the weight parameter $\lambda$ in $\mathcal{L}$ to be 2 for roughly scaling the gradient magnitudes of $\mathcal{L}_{\text{VTC}}$ and $\mathcal{L}_{\text{TS}}$ to be the same.
For the text encoder of all models, we freeze the first three-quarters of layers, leaving other layers trainable.
Due to computation source limitation, we vary the batch size for different models, and all hyper-parameters are listed in Table~\ref{tab:hyper_params}.

\section{Divided Space-Time Attention}

\begin{figure}[t]
    \centering
    \includegraphics[width=.8\textwidth]{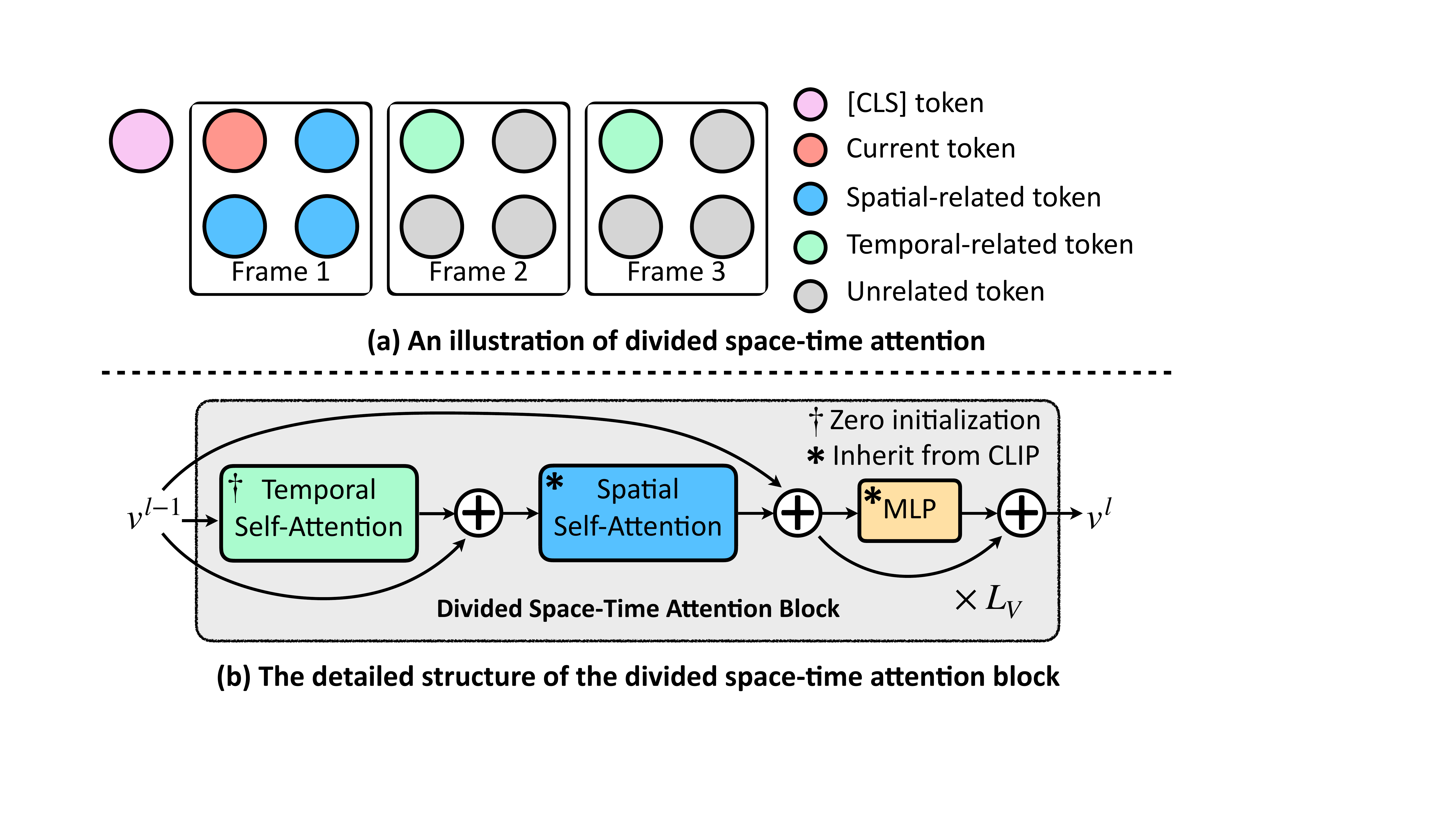}
    \caption{(a) We illustrate the divided space-time attention. For the orange token, it attends to tokens belonging to the same frame, \ie spatial-related tokens, and tokens in the same position across different frames, \ie temporal-related tokens, as well as the [CLS] token. (b) The structure of the divided space-time attention block, where we initialize the parameters of the temporal self-attention module with zeros and inherit weights from CLIP for the spatial self-attention and MLP modules.}	
\label{fig:supp_attn}
\end{figure}

In this section, we describe the divided space-time attention in our video encoder detailedly.
As illustrated in Figure~\ref{fig:supp_attn}, for the intra-frame tokens, \ie spatial-related tokens, we add the same spatial positional embeddings, and for tokens at the same position across different frames, \ie temporal-related tokens, we add the same temporal positional tokens. 
For each token, it first attends to the temporal-related tokens, then attend to the spatial-related tokens.
Note that the [CLS] token is attended in both temporal and spatial self-attention.

\section{Additional Experiments}

\vspace{-0.5em}
\noindent

\begin{wraptable}{r}{5cm}
\vspace{-1.5em}
\caption{The R@1 \wrt different temporal initialization strategies.}
\centering
\resizebox{.35\textwidth}{!}{
\begin{tabular}{ccc}
\toprule
Temp Init & MSR-VTT & DiDeMo \\
\midrule
zero      & \textbf{34.5}    & \textbf{31.5}   \\
random    & 31.2    & 29.3   \\
\bottomrule
\end{tabular}}
\label{tab:ab_supp}
\end{wraptable}
\noindent \textbf{Temporal Initialization.}
We analyze the proper way to instantiate temporal modules based on the ViT-B/32 model.
As shown in Table~\ref{tab:ab_supp}, initializing temporal attention weights with zeros beats its random initialization competitor by a large margin, indicating that growing temporal reasoning ability out of nothing avoids hurting the well-learned spatial prior, thus more suitable for training video foundation models.

\begin{figure}[t]
    \centering
    \includegraphics[width=.7\textwidth]{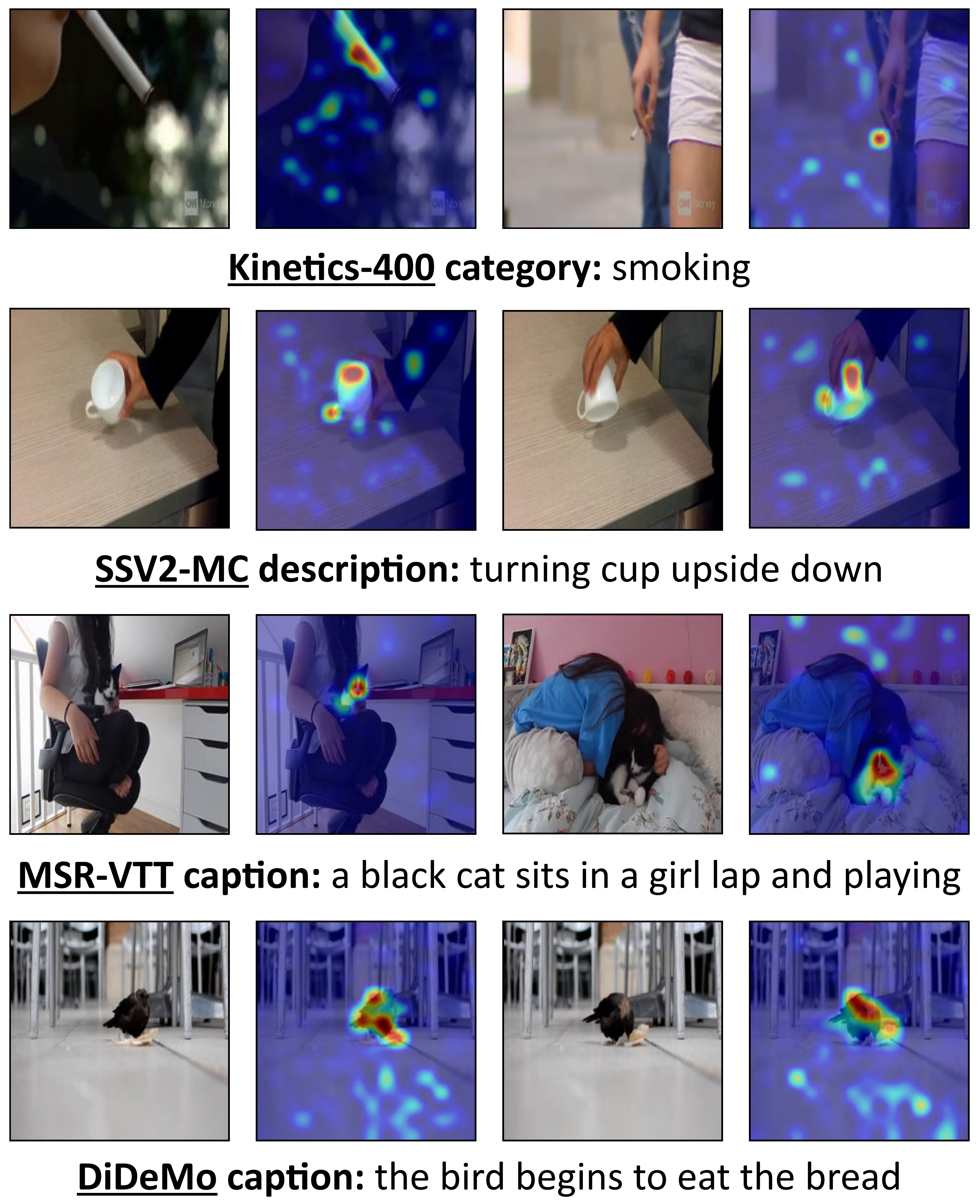}
    \caption{The self-attention map of the visual [CLS] token. \emph{Note that it does not involve cross-modal alignment.} The captions are for reference only and are not used for attention maps. All crucial objects involved in the temporal interaction are captured precisely, which indicates our model's powerful ability to produce general-purpose features.}	
\label{fig:supp_vis}
\end{figure}

\noindent \textbf{Attention Visualization.}
We visualize the zero-shot self-attention map of the visual [CLS] token on different datasets in Figure~\ref{fig:supp_vis}.
The critical objects involved in the temporal interactions are captured precisely, indicating our model's strong out-of-the-box ability intuitively.

\section{Broader Impact}

The negative social impacts of our model may lie in intensifying global warming because of the large amount of carbon emission produced by GPU clusters.
Though the pre-training phase is energy-consuming, the model can be used out-of-the-box, saving the potential carbon emission in fine-tuning.
Since the model produces general-purpose representations, it might also raise the risk of abuse, such as unauthorized biometric recognition.

\begin{table}[t]
\setlength{\tabcolsep}{2pt}
\centering
\caption{Our detailed model architectures. ``Embed'' and ``Hidden'' denote the dimension of the shared and hidden representations, respectively.}
\resizebox{.95\textwidth}{!}{
\begin{tabular}{ccccccccccccc}
\toprule
\multirow{2}{*}{Model} & \multirow{2}{*}{Params} & \multirow{2}{*}{Embed} & \multirow{2}{*}{GFLOPs} & \multicolumn{5}{c}{Video  Encoder}        & \multicolumn{4}{c}{Text Encoder}  \\
\cmidrule{5-13}
                       &                         &                        &                         & Layers & Params & Hidden & Patch & GFLOPs & Layers & Params & Hidden & GFLOPs \\
\midrule
B/32                   & 149M                    & 512                    & 72                      & 12     & 86M    & 768    & $32\times 32$    & 69     & 12     & 63M    & 512    & 3      \\
B/16                   & 149M                    & 512                    & 281                     & 12     & 86M    & 768    & $16\times 16$    & 278    & 12     & 63M    & 512    & 3      \\
H/14                   & 1.0B                    & 1024                   & 2674                    & 32     & 632M   & 1280   & $14\times 14$    & 2650   & 24     & 354M   & 1024   & 24    \\
\bottomrule
\end{tabular}}
\label{tab:model_arch}
\end{table}

\begin{table}[t]
\centering
\caption{The hyper-parameters for pre-training and liner probe.}
\resizebox{.75\textwidth}{!}{
\begin{tabular}{lcccc}
\toprule
\multirow{2}{*}{config}        & \multicolumn{3}{c}{pre-training}        & linear probe         \\
\cmidrule {2-5}
                               & B/32          & B/16        & H/14      & all models           \\
\midrule
optimizer                      & \multicolumn{3}{c}{AdamW}               & SGD                  \\
weight decay                   & \multicolumn{3}{c}{0.05}                & 0                    \\
training epochs                & \multicolumn{2}{c}{10}      & 5         & 100                  \\
\multirow{2}{*}{learning rate} & \multicolumn{3}{c}{$1\times 10^{-4}$ (new modules)}  & \multirow{2}{*}{0.1} \\
                               & \multicolumn{3}{c}{\phantom{p}$1\times 10^{-7}$ (CLIP modules)} &                      \\
batch size                     & 768           & 768         & 160       & 512                  \\
frozen text layers             & \multicolumn{2}{c}{1-9}     & 1-18      & n/a                  \\
input frames                   & \multicolumn{3}{c}{12}                  & 12                   \\
masking ratio                  & 0             & 50          & 70        & 0                    \\
augmentation              & \multicolumn{3}{c}{RandomCrop}          & CenterCrop           \\
GPU for training               & 32 V100       & 64 V100     & 80 V100   & 16 V100          \\
\bottomrule
\end{tabular}}
\label{tab:hyper_params}
\end{table}

\end{document}